\documentclass{article} % For LaTeX2e
\usepackage{iclr2023_conference,times}

\usepackage{amsmath,amsfonts,bm}

\def\eqref#1{equation~\ref{#1}}
\def\1{\bm{1}}

\DeclareMathAlphabet{\mathsfit}{\encodingdefault}{\sfdefault}{m}{sl}
\SetMathAlphabet{\mathsfit}{bold}{\encodingdefault}{\sfdefault}{bx}{n}

\usepackage{hyperref}
\usepackage{url}
\usepackage{tipa}
\usepackage{booktabs}
\usepackage{subcaption}
\usepackage{comment}
\usepackage{xspace}

\DeclareUnicodeCharacter{1E13}{\d}
\newcommand*{\mphaya}{MphayaNER\xspace}
\newcommand*{\venda}{Tshiven\textsubcircum{d}a\xspace}

\title{MphayaNER: Named Entity Recognition for Tshiven\textsubcircum{d}a}

\author{
Rendani Mbuvha\textsuperscript{*,1,2}, David I. Adelani\textsuperscript{3}, Tendani Mutavhatsindi\textsuperscript{2}, Tshimangadzo Rakhuhu\textsuperscript{4}, \\
\textbf{Aluwani Mauda\textsuperscript{5}, Tshifhiwa Joshua Maumela\textsuperscript{6}, Andisani Masindi,
Seani Rananga\textsuperscript{7},} \\
\textbf{Vukosi Marivate\textsuperscript{7,8}, and Tshilidzi Marwala\textsuperscript{9}} \\
\textsuperscript{1} School of Electronic Engineering and Computer Science, Queen Mary University of London.  \\
\textsuperscript{2} School of Statistics and Actuarial Science, University of Witwatersrand  \\
\textsuperscript{3} Department of Computer Science, University College London. \\
\textsuperscript{4} School of Business,  National College of Ireland  \\
\textsuperscript{5} Centre for Development Support, University of the Free State  \\ 
\textsuperscript{6} School of Electrical and Electronic Engineering, University of Johannesburg \\
\textsuperscript{7} Department of Computer Science, University of Pretoria  \\
\textsuperscript{8} Lelapa AI  \\
\textsuperscript{9} United Nations University  \\
\textsuperscript{*} Corresponding email \texttt{r.mbuvha@qmul.ac.uk}
}
\newcommand{\scd}{\textsubcircum{d}}

 \iclrfinalcopy % Uncomment for camera-ready version, but NOT for submission.
\begin{document}

\maketitle
\begin{abstract}
Named Entity Recognition (NER) plays a vital role in various Natural Language Processing tasks such as information retrieval, text classification, and question answering. However, NER can be challenging, especially in low-resource languages with limited annotated datasets and tools. This paper adds to the effort of addressing these challenges by introducing MphayaNER, the first Tshiven\textsubcircum{d}a NER  corpus in the news domain. We establish NER baselines by \textit{fine-tuning} state-of-the-art models on MphayaNER. The study also explores zero-shot transfer between Tshiven\textsubcircum{d}a and other related Bantu languages, with chiShona and Kiswahili showing the best results. Augmenting MphayaNER with chiShona data was also found to improve model performance significantly. Both MphayaNER and the baseline models are made publicly available~\footnote{\url{https://github.com/rendanim/MphayaNER}}.
\end{abstract}
Named Entity Recognition (NER) is a natural language processing (NLP) task that involves identifying and classifying named entities such as person names, location names, and organization names in text \citep{Mohit2014,eiselen2016government}. NER is critical in various NLP applications, including information retrieval, text classification, and question answering. Identifying named entities is challenging, particularly in low-resource languages where annotated datasets and tools for NER are limited \citep{adelani2021masakhaner}. Tshiven\textsubcircum{d}a is a minority language spoken by approximately 1.3 million\footnote{\url{http://salanguages.com/Tshivenda/index.htm}} people, predominantly in South Africa and its northern neighbours of Zimbabwe, Mozambique and Botswana \cite{africa}. Given the relatively small number of native speakers of the language, Tshiven\textsubcircum{d}a has received limited attention in the NLP community, especially concerning NER. NER activity in Tshiven\textsubcircum{d}a has been limited to SADiLaR corpus, which is primarily restricted to named entities in the government domain \citep{eiselen2016government}. The task of NER in Tshiven\textsubcircum{d}a is complicated by the fact that Tshiven\textsubcircum{d}a, like many other Bantu languages, is a morphologically rich language with complex morphological forms that can pose challenges for NER systems \citep{van2014translation,adelani-etal-2022-masakhaner}. Additionally, Tshiven\textsubcircum{d}a has limited resources, including annotated datasets and pre-trained models, which are essential for developing NER systems. These challenges highlight the need to develop NER systems tailored to Tshiven\textsubcircum{d}a, specifically designed to handle the language's morphological richness and resource-poor nature. We address the underrepresentation of Tshiven\textsubcircum{d}a in NER  by \textit{fine-tuning} state-of-the-art pre-trained language models and further investigate cross-lingual zero-shot transfer between Tshiven\scd a and five other African languages.

\subsection*{ Tshiven\textsubcircum{d}a}

Tshiven\textsubcircum{d}a is part of the South-Eastern group of Bantu languages and is known for its similarities to the Karanga Dialect of Shona, as noted in \citep{cassimjee1992autosegmental}. Tshiven\textsubcircum{d}a employs the Latin alphabet along with five letters that have accents. Four of these letters are dental consonants with a circumflex accent beneath them (\textsubcircum{d}, \textsubcircum{l}, \textsubcircum{n}, \textsubcircum{t}). One is a dot above the letter for the velar ($\dot{\text{n}}$). Seven vowels are represented using five vowel letters. J and Q are used exclusively for foreign words and names. Tshiven\textsubcircum{d}a has a rich morphological structure featuring 21 noun classes \citep{musehane_2014}. Like many Bantu languages, it only uses two tones - high (denoted by an acute accent) and low (no diacritic) \citep{reynolds1997post}. The population of Tshiven\textsubcircum{d}a speakers is the second least in South Africa after isiNdebele speakers. As part of the minority speakers, there has been little attention to ensuring this language is included in digital platforms, particularly news media.

\subsection*{MphayaNER Corpus}
We use data from Vuk\textquotesingle uzenzele \footnote{\url{https://www.vukuzenzele.gov.za/}}, a monthly newspaper published in multiple languages, including Tshiven\scd a, by the communications unit of the South African Government. For this study, Tshiven\textsubcircum{d}a Vuk\textquotesingle uzenzele news articles were used. The articles were originally in PDF format, they were converted to text files using py2PDF \citep{pypdf2}, cleaned and standardised by the Data Science for social impact (DSFSI) research group at the University of Pretoria \cite{marivate_vukosi_2023_7598540,lastrucci2023preparing}\footnote{\url{https://github.com/dsfsi/vukuzenzele-nlp/}}. We manually preprocess the converted text file to minimise text extraction errors such as repeated phrases and erroneous word concatenation or splitting to create the MphayaNER corpus.  

\subsection*{Annotation} We make use of the same annotation guideline, scheme and tags as done by \citep{adelani2021masakhaner}. We also make use of ELISA annotation tool for labelling Tshiven\textsubcircum{d}a texts. We recruited 5 native speakers for the labelling and they were trained using the MasakhaNER guideline for labeling personal name (PER), organization (ORG), location (LOC), and dates (DATE). To speed up annotation effort, we assigned a sentence to be annotated by only two annotators. After annotation, we make use of the adjudication function of the ELISA tool to detect mistakes between annotators. Annotators met to resolve conflict, and for sentences that still had disagreement, we ask the lead Tshiven\textsubcircum{d}a annotator to decide the most appropriate tag for the token. Our new dataset is known as MphayaNER. After annotation and verification, we divided the 1,359 sentences into TRAIN, DEV and TEST split based on 70\%/10\%/20\% ratio. In total, the dataset consists of 40,778 tokens, where TRAIN, DEV and TEST splits have about 29,239, 4,212 and 7,327 tokens respectively as shown in \autoref{tab:results}. In MphayaNER, there are few entities in general, we found only 251 PER, 409 LOC, 702 ORG , and 504 DATE entities.  

\subsection*{Baseline Models and Results} We provide baseline NER models based on \textit{fine-tuning} multilingual pre-trained language models (PLMs) -- this is the state-of-the-art technique. We compared fine-tuning mDeBERTaV3~\citep{he2021debertav3}, one of the current state-of-the-art PLMs and two Africa-centric PLMs, i.e. AfriBERTa\citep{ogueji-etal-2021-small} and AfroXLMR~\cite{alabi-etal-2022-adapting}. We benchmark on several sizes of the AfroXLMR model (small, base, and large versions). All models are trained for 20 epochs, a learning rate of $5e-5$. 

Table 1(a) shows the result on MphayaNER; the AfriBERTa model has the worst result ($58.7$ F1) since Tshiven\textsubcircum{d}a and other related Southern Bantu languages were not covered during pre-training. mDeBERTaV3 had an impressive result ($66.4$ F1) despite seeing only two Bantu languages during pre-training(Kiswahili and isiXhosa). AfroXLMR has seen more Bantu languages, about eight. We found that the more the capacity/size of AfroXLMR increases, the better the performance --- AfroXLMR-small ($66.1$ F1), AfroXLMR-small ($67.3$ F1), and AfroXLMR-large ($71.0$ F1). We found all the models struggle to identify DATE in the corpus. 

\vspace{-2mm}
\paragraph{Transfer result} We examine how other Bantu languages adapt to  \venda in the zero-shot setting. Similarly, we compared the adaptation of a NER model trained on several Southern African languages in MasakhaNER 2.0 corpus~\citep{adelani-etal-2022-masakhaner} t\venda. The source transfer languages are chiShona (sna), Kiswahili(swa), Setswana(tsn), isiXhosa (xho), and isiZulu (zul). Also, we perform a zero-shot adaptation of SADILAR \venda NER corpus based on government data to \mphaya corpus based on the news corpus (\textit{govt $\rightarrow$ news}). We excluded the miscellaneous tag (MISC) and DATE tag for the cross-domain experiment since the \mphaya dataset does not have MISC.

\autoref{tab:tranfer_results} shows the result of the cross-lingual transfer with or without the DATE tag. Our result in Table 2(a) shows a significant drop in performance when transferring to \venda for all languages. Surprisingly, the government domain transfer ($35.2$ F1) to \mphaya is even worst than language adaptation from other Southern African languages ($>37.0 F1$), which shows that the difference in domain  has a big effect on transfer performance. Furthermore, we performed transfer experiments, including the DATE tags for the top three languages with the best transfer results in Table 2(a). We found out that Kiswahili and chiShona have very good transfer results for the DATE tag (over $64.0$ F1); the performance for Kiswahili was particularly unexpected since it is not a Southern Bantu language like others. Setswana (tsn) seems to be better on LOC entity since native speakers of the languages are geographically close (i.e both are spoken in the same region in the northeastern parts of South Africa). All the source languages struggle to adapt to both ORG and PER entities. Lastly, we performed co-training experiments by combining the \mphaya corpus with the one of chiShona, Kiswahili and Setswana; we found that co-training improves the performance by $2.5-3.9$ F1 points. In general, only co-training with Setswana leads to $3.5$ F1 points. %While the original \mphaya baseline struggle with DATE tagging, co-training significantly improves transfer to DATE entities. This result highlights the importance of increasing the training data of MphayaNER with sentences with more entities. We leave this as future work.  

%Tshiven\textsubcircum{d}a. The best result ($>52.0$ F1) was achieved by chiShona and Kiswahili. On further analysis, we found both languages are better able to identify DATE than MphayaNER training corpus reaching $71.0$ F1 for DATE. \textbf{As a last experiment}, we combined \textbf{chiShona data} from M2.0 with \textbf{MphayaNER}, and we obtained a better result of ($75.3$) especially for DATE ($47.6 \rightarrow 56.0$) F1. This result highlights the importance of increasing the training data of MphayaNER with sentences with more entities. We leave this as future work.  

\begin{table}[t]
    \footnotesize
    \begin{subtable}{0.35\linewidth}
      \centering
        \begin{tabular}{lrr}
        \toprule
        & \textbf{\# of } & \textbf{\# of}  \\
        \textbf{Split} & \textbf{sentences} & \textbf{tokens}  \\
        \midrule
        TRAIN & 951 & 29,239 \\
        DEV & 135 & 4,212 \\
        TEST & 273 & 7,327 \\
        \bottomrule
        \end{tabular}
        \vspace{0.4mm}
        \label{tab:data_split}
        \caption{Data split for MphayaNER}
    \end{subtable} %
    \begin{subtable}{0.7\linewidth}
      \centering
      \setlength\tabcolsep{4pt} %default 6pt
        \begin{tabular}{p{26mm}rrrr|r}
        \toprule
         & \multicolumn{5}{c}{\textbf{F1-score}}\\
        \textbf{Model} & \textbf{DATE} & \textbf{LOC} & \textbf{ORG}  & \textbf{PER} & \textbf{AVG}\\
        \midrule
        mDeBERTaV3 & 39.2 & 76.6 & 67.6 & 96.2 & $66.4$\\
        AfriBERTa & 37.4 & 60.2 & 54.8 & 84.2 & $58.7$\\
        AfroXLMR-small & 41.0 & 77.4& 62.4 & \textbf{99.4} & $66.1$\\
        AfroXLMR-base & 43.4 & \textbf{80.2} & 62.6  & 97.0  &  $67.3$\\
        AfroXLMR-large & \textbf{47.6} &  77.2 & \textbf{71.0} & 98.8 & $\mathbf{71.0}$\\
        
        \bottomrule
        \end{tabular}
        \label{tab:baseline}
        \caption{Baseline results for MphayaNER.  Average over 5 runs. }
    \end{subtable}
    \vspace{-5mm}
    \caption{Data split and Benchmark results (F1-score) for MphayaNER. }
    \label{tab:results}
\end{table}

\begin{table}[t]
    \footnotesize
    \begin{subtable}{0.35\linewidth}
      \centering
      \setlength\tabcolsep{5.pt} %default 6pt
       \begin{tabular}{lrrr|r}
        \toprule
         & \multicolumn{4}{c}{\textbf{F1-score}}\\
        \textbf{lang.} & \textbf{LOC} & \textbf{ORG}  & \textbf{PER} & \textbf{AVG}\\
        \midrule
        \multicolumn{3}{l}{\textbf{\textit{src lang. $\rightarrow$ tgt lang.}}} \\
        %nya  & 32.0 & 28.6 & 38.6 & $31.9$\\
        sna  & 47.6 & 30.6 & \textbf{46.0} & $40.0$\\
        swa  &  52.4 & \textbf{37.2}  & 40.6 & $43.2$\\
        tsn  & \textbf{66.4} & 31.4 & 45.0 & $\mathbf{46.2}$\\
        xho  & 57.0 & 29.3 & 35.3 & $38.4$\\
        zul  & 50.8 & 29.8 & 32.6 & $37.0$\\
        \midrule
        \multicolumn{4}{l}{\textbf{same lang.: \textit{govt $\rightarrow$ news}}} \\
        ven  & 53.4  & 33.6 & 20.8  & $35.2$ \\
        \bottomrule
        \end{tabular}
        \label{tab:zero-shot-eval}
        
        \caption{Zero-shot eval. on MphayaNER}
    \end{subtable} %
    \begin{subtable}{0.7\linewidth}
      \centering
      \setlength\tabcolsep{4pt} %default 6pt
        \begin{tabular}{lrrrr|r}
        \toprule
         & \multicolumn{5}{c}{\textbf{F1-score}}\\
        \textbf{src lang.} & \textbf{DATE} & \textbf{LOC} & \textbf{ORG}  & \textbf{PER} & \textbf{AVG}\\
        \midrule
        %M2.0 & \textbf{63.0} &  61.0 & 26.0 & 43.0 & $\mathbf{46.1}$\\
        \multicolumn{3}{l}{\textbf{\textit{zero-shot eval.}}} \\
        sna & 64.6 & 49.2 & 30.2 & 38.8 & $42.9$\\
        swa & \textbf{72.6} & 32.6 & 37.4 & 38.8 &  $43.2$\\
        tsn & 56.0 & 52.4 & 29.6 & 39.4 & $43.7$\\
        \midrule
        \multicolumn{3}{l}{\textbf{\textit{co-training eval. }}} \\
        sna+ven & 50.0 &  82.4 & 72.8 & 97.0 & $73.5$\\
        swa+ven & 48.0 & 81.2 & 75.4 & \textbf{97.4} & $73.5$\\
        tsn+ven & 50.0 & \textbf{86.2} & \textbf{76.8} & 94.6 & $74.5$\\
        sna+swa+tsn+ven & 52.6 & 86.0 & 75.6 & 93.0 & $\mathbf{74.9}$\\
        \bottomrule
        \end{tabular}
        \label{tab:baseline}

        \caption{zero-shot and co-training results on all named entity tags}
    \end{subtable}

    \caption{Transfer learning and co-training evaluation results on MphayaNER. }
    \label{tab:tranfer_results}
\end{table}

 \vspace{-1mm}
 
 \subsection*{Conclusion}
 \vspace{-2mm}
 In this paper, we present first Tshiven\textsubcircum{d}a NER corpus in the news domain based on the Vuk\textquotesingle zenzele dataset. We establish baselines for this task by fine-tuning multilingual pre-trained state-of-the-art language models. We also investigate zero-shot transfer between Tshiven\textsubcircum{d}a and related Bantu languages finding that Setswana and Kiswahili give the best results. Additional experiments show augmenting MphayaNER with chiShona data significantly improves model performance.

\bibliography{iclr2023_conference}
\bibliographystyle{iclr2023_conference}

%\appendix
%\section{Appendix}
%You may include other additional sections here.

\end{document}